\documentclass[conference]{IEEEtran}

\usepackage{hang}
\usepackage{times}
\usepackage{comment}

\usepackage[numbers]{natbib}
\usepackage{multicol}
\usepackage[hidelinks,bookmarks=true,colorlinks=false]{hyperref}

\IEEEoverridecommandlockouts 

\usepackage{subcaption}

\usepackage{graphicx}
\usepackage{overpic}
\usepackage{xcolor}
\usepackage{tikz}

\usepackage{amsmath, mathtools, amssymb, amsthm}
\usepackage{nicefrac}
\usepackage{mathrsfs}
\usepackage{dsfont}
\usepackage{siunitx}
\usepackage{caption}
\usepackage{cuted}

\usepackage{algorithm}
\usepackage[noend]{algpseudocode}

\usepackage[capitalize,nameinlink]{cleveref}
\crefformat{equation}{(#2#1#3)}
\usepackage[acronyms, shortcuts, hyperfirst=false]{glossaries}
\glsdisablehyper

\newcommand{\norm}[1]{\mathord{\left\lVert#1\right\rVert}}

\newtheorem{remark}{Remark}
\newtheorem{assumption}{Assumption}

\newacronym{rl}{RL}{reinforcement learning}
\newacronym{mppi}{MPPI}{Model Predictive Path Integral}
\newacronym{mpc}{MPC}{Model Predictive Control}
\newacronym{pomdp}{POMDP}{partially observable Markov decision process}

\newcommand{\Method}{\texttt{SimDist}}

\newcommand{\dataset}{\mathcal{D}}
\newcommand{\simdataset}{\dataset_\mathtt{sim}}
\newcommand{\realdataset}{\dataset_\mathtt{real}}
\newcommand{\eps}{\varepsilon}

\newcommand{\simstate}{s}

\newcommand{\proprio}{o}
\newcommand{\action}{a}

\newcommand{\latent}{z}

\newcommand{\reward}{r}
\newcommand{\val}{v}
\newcommand{\expertflag}{b^e}

\newcommand{\contextencoder}{C}
\newcommand{\encoder}{E}
\newcommand{\dynamicsmodel}{f}
\newcommand{\rewardmodel}{R}
\newcommand{\valuemodel}{V}
\newcommand{\bcpolicy}{\pi}

\newcommand{\panel}[3]{
  \begin{overpic}[width=#3]{#1}
    \put(0.9,5){%
      \tikz[baseline]{
        \node[
          fill=black,
          fill opacity=0.5,
          text=white,
          text opacity=1,
          rounded corners=2pt,
          inner xsep=4pt,
          inner ysep=2pt
        ]{\bfseries #2};
      }
    }
  \end{overpic}
}

\begin{document}

\title{Simulation Distillation: Pretraining World Models in Simulation for Rapid Real-World Adaptation
\vspace{-0.25em}
}

\author{
  Jacob Levy$^{\ast1}$\quad Tyler Westenbroek$^{\ast2}$\quad Kevin Huang$^{2}$\quad Fernando Palafox$^{1}$\quad
  Patrick Yin$^2$\quad\\
  Shayegan Omidshafiei$^3$\quad Dong-Ki Kim$^3$\quad Abhishek Gupta$^{\dagger2}$\quad David Fridovich-Keil$^{\dagger1}$\\
  $^1$University of Texas at Austin\quad $^2$University of Washington\quad $^3$FieldAI \\
  \texttt{\textcolor{magenta}{\url{https://sim-dist.github.io}}}
\vspace{-1.75em}
}

\maketitle

\begingroup
\setlength\stripsep{0pt}
\begin{strip}
    \centering
    \includegraphics[width=.99\textwidth]{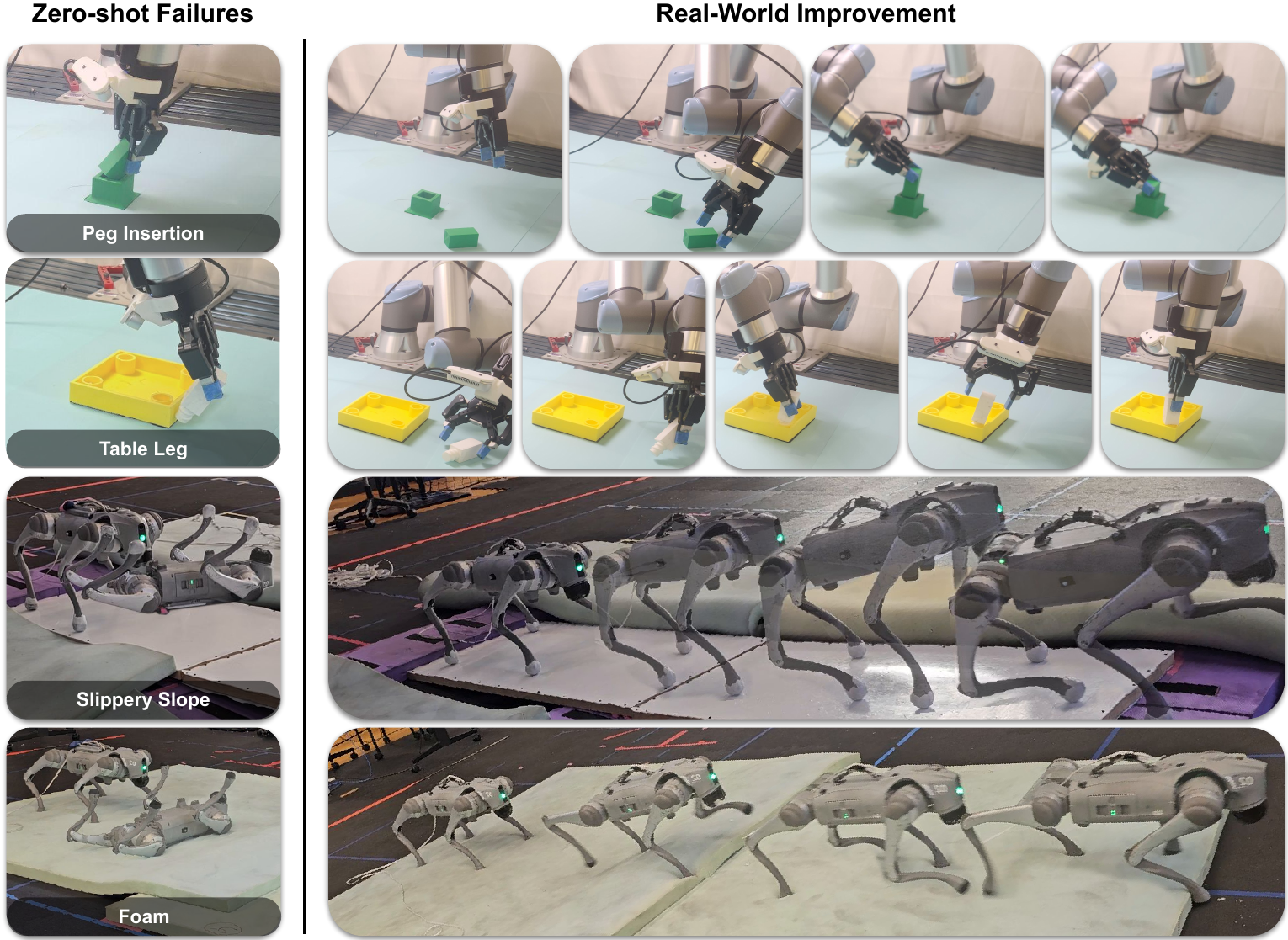}
    \captionof{figure}{\footnotesize{ Failures of zero-shot sim-to-real policies  (left). Our framework \Method\ rapidly overcomes the dynamics gap and improves performance with minimal real-world interaction. We demonstrate substantial gains in task execution on both precise manipulation and quadrupedal locomotion tasks with only 15-30 minutes of real-world data, substantially outperforming baselines. 
    }}
    \label{fig:front}
    \vspace{1em}
\end{strip}
\endgroup

\begin{abstract}
Robot learning requires adaptation methods that improve reliably from limited, mixed-quality interaction data. This is especially challenging in long-horizon, contact-rich tasks, where end-to-end policy finetuning remains inefficient and brittle. World models offer a compelling alternative: by predicting the outcomes of candidate action sequences, they enable online planning through counterfactual reasoning. However, training action-conditioned robotic world models directly in the real world requires diverse data at impractical scale. We introduce \texttt{Simulation Distillation} (\Method), a framework that uses physics simulators as a scalable source of action-conditioned robot experience. During pretraining, \Method\ distills structural priors from the simulator into a world model that enables planning from raw real-world observations. During real-world adaptation, \Method\ transfers the encoder, reward model, and value function learned in simulation, and updates only the latent dynamics model using real-world prediction losses. This reduces adaptation to supervised system identification while preserving dense, long-horizon planning signals for online improvement. Across contact-rich manipulation and quadruped locomotion tasks, \Method\ rapidly improves with experience, while prior adaptation methods struggle to make progress or degrade during online finetuning. Project website and code: \texttt{\textcolor{blue}{\url{https://sim-dist.github.io}}}

$^\ast$Equal contribution,  $^\dagger$Equal advising
\end{abstract}

\IEEEpeerreviewmaketitle

\begin{figure*}[t]
    \centering
    \includegraphics[width=.99\textwidth]{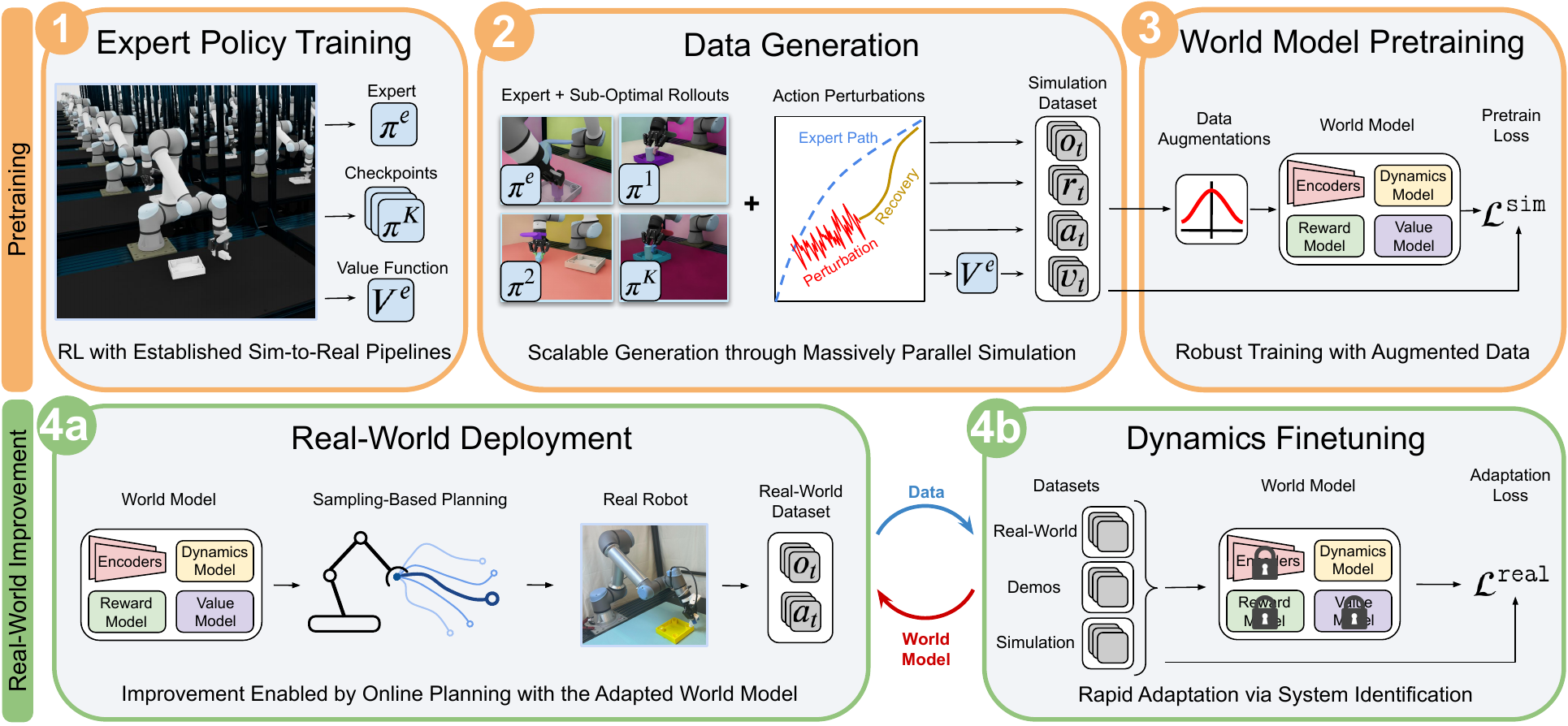}
    \caption{
    \footnotesize{\Method \ overview.
    1) An expert policy, policy checkpoints, and a value function are trained in simulation using privileged state. 2) Large-scale training data are generated by combining expert and sub-optimal policies with contiguous action perturbations, yielding diverse trajectories with dense reward and value supervision. 3) A planning-oriented latent world model is pretrained on this data, learning representations, dynamics, rewards, and values from raw observations. 4a) At deployment, the learned representation and dense reward and value models are transferred to the real robot to enable planning with the latent dynamics. 4b) Real-world data is then used to finetune only the dynamics via supervised system identification, with representations, rewards, and values frozen. Deployment and finetuning are iterated, enabling rapid and stable real-world adaptation.
    }
    }
     \vspace{-1em}
    \label{fig:method}
\end{figure*}

\section{Introduction}

Robotic systems must effectively adapt their behavior using limited interactions in new environments. This adaptation data is often of mixed quality, spanning demonstrations, failed attempts, exploratory actions, and rollouts from previous policies. An effective adaptation algorithm should preserve useful priors from pretraining while extracting maximal information from each new in-domain sample. This is especially important in long-horizon, contact-rich tasks, where small errors compound and success requires reasoning through many possible futures.

We argue that world models \cite{bruce2024genie, zhu2025unified}, rather than monolithic end-to-end policies \cite{intelligence2025pi, intelligence2025pi05, kim2024openvla}, provide the right abstraction for leveraging prior experience to improve efficiently in new environments. Existing end-to-end reinforcement learning methods \cite{rlpd, haarnoja2018learning} often collapse when finetuning policies in new domains \cite{yin2025rapidly, smith2022legged, smith2022walk}, indicating catastrophic forgetting of pretraining priors \cite{zhou2024efficient}. This reflects a limitation of off-policy model-free finetuning: task representations, reward and value estimates, dynamics, and action selection are tightly entangled, so adaptation updates the entire decision-making process end-to-end while solving difficult long-horizon credit assignment problems. In contrast, world model architectures \cite{hafnerdream, hansen2022temporal} typically \emph{modularize decision making} by learning separate networks for environment prediction and credit assignment. This separation allows new environment data to refine the model of action consequences without overwriting the broader decision-making structure learned during pretraining. Online planning can then convert these improved predictions into better behavior by evaluating counterfactual futures beyond those directly observed in the robot's data.

However, learning a world model suitable for planning requires action-conditioned robot data with diverse coverage at a scale that is prohibitively expensive to collect purely in the real world. Planning algorithms \cite{williams2016aggressive} sample candidate trajectories beyond the optimal data distribution and must \emph{discriminate} action sequences that lead to success from those that lead to failure over long horizons. This requires dynamics predictions, return estimates, and state representations that generalize over mistakes, corrections, and varying contact sequences. \emph{How can we obtain the data coverage and supervision required to train these models at scale?}

We argue that simulation, despite the sim-to-real gap, provides an ideal setting for bootstrapping the components of a world model that are required for effective real-world decision making. Beyond cheap, scalable interaction data, privileged simulator state enables supervision that is difficult to obtain at scale in the real world. Simulator rewards can train dense reward models that provide informative planning signals from raw perception. Expert policies and critics learned by existing student-teacher RL pipelines \cite{yin2025rapidly, lee2020learning} provide scalable labels for action priors and long-horizon value estimation. Together, these rich sources of supervision encourage the world-model encoder to learn robust state representations that capture the features required for effective decision making \cite{hansen2022temporal}. However, a natural concern remains: if a simulator does not exactly replicate the real world, won't a world model inherit the same biases, leading to poor real-world performance?

Our key insight is that these components need not be perfectly calibrated to the real world to enable effective planning during deployment. Indeed, reward and value models only need to define an accurate \emph{ranking} over real-world states to enable the planner to distinguish promising futures from poor ones. This is a weaker and more transferable requirement than estimating exact returns \cite{yin2025rapidly}. For example, in our peg insertion task, the model need only rank states where the peg is closer to the hole, better aligned, or partially inserted above states farther from success. The dynamics model, in contrast, ties actions to future states and is particularly sensitive to the dynamics gap. However, with a properly initialized model, this gap can be corrected efficiently using simple, supervised finetuning on real-world data. 

Concretely, we introduce \texttt{Simulation Distillation} (\Method), a framework for bootstrapping world models in simulation and efficiently adapting in the real world with online planning and dynamics adaptation. During simulation pretraining, we systematically generate diverse rollouts by perturbing expert action sequences to expose the model to mistakes, recoveries, and failed attempts. At deployment, we freeze the state encoder, reward model, and value function learned in simulation, and update only the latent dynamics model using real-world prediction losses. The frozen reward and value heads provide immediate long-horizon planning signals, enabling a relatively short-horizon planner to improve performance as dynamics predictions become more accurate. We provide extensive ablations showing that the encoder, reward model, and value function transfer robustly across the sim-to-real gap, and that broad simulation pretraining enables the dynamics model to generalize to new trajectories from limited real-world data. Altogether, \Method\ sidesteps the long-horizon credit assignment and bootstrapping problems that make real-world RL unstable and data-hungry by moving this bootstrapping to simulation. Across four real-world tasks spanning quadruped locomotion and contact-rich manipulation, \Method\ reliably improves with additional experience, while baseline methods for online finetuning struggle to make meaningful progress and can even degrade during adaptation.

We summarize our contributions as follows:

\begin{hangingpar}
\textbf{1)} We introduce \Method, a world-model framework for sim-to-real transfer that reduces real-world adaptation to supervised dynamics learning while reusing reward, value, and representation priors learned in simulation.
\end{hangingpar}

\begin{hangingpar}
\textbf{2)} We instantiate \Method\ on two contact-rich manipulation tasks and two quadruped locomotion tasks in the real world, achieving reliable autonomous improvement with only 15--30 minutes of real-world data and substantially outperforming existing adaptation strategies. 
\end{hangingpar}
\section{Related Work}

\begin{figure}[tb]
    \centering
    \includegraphics[width=.48\textwidth]{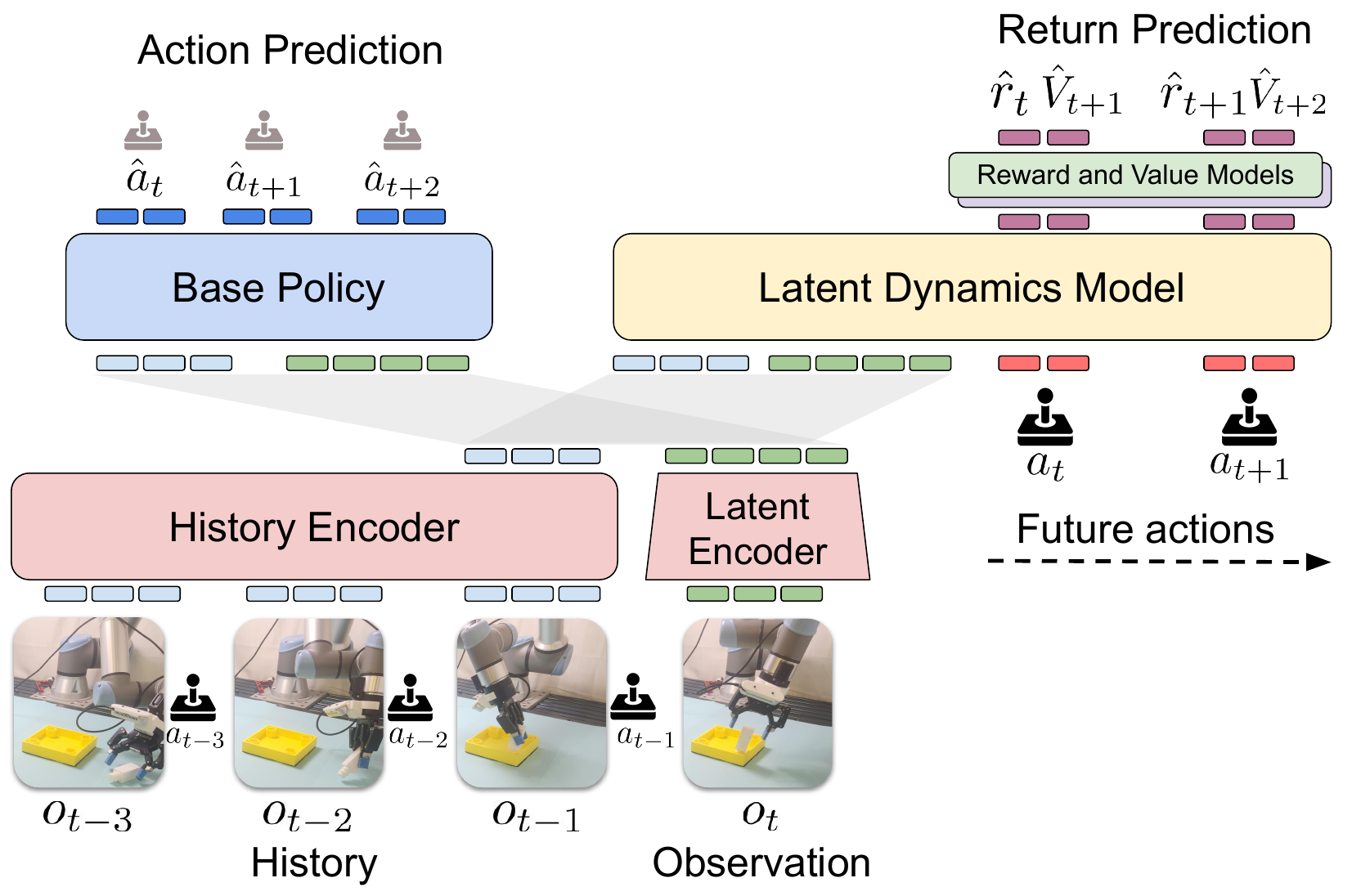}
    \caption{
    \footnotesize{
    World model architecture.
    The most recent observation is encoded into a latent representation while a history encoder processes a history of observations and actions. These jointly condition a transformer-based latent dynamics model that predicts future latent trajectories under candidate action sequences. Transformer-based reward and value heads evaluate predicted trajectories to produce reward and value sequences, while a base policy head predicts action chunks used to warm-start sampling-based planning.}
    }
    \vspace{-1em}
    \label{fig:model}
\end{figure}

\textbf{Real-World Reinforcement Learning.} A growing body of work studies reinforcement learning on real-world robotic systems \cite{serl, levy2025learning, yin2025rapidly, smith2022walk, westenbroek2022lyapunovdesignrobustefficient, westenbroek2023enabling, JMLR:v17:15-522, pmlr-v87-kalashnikov18a, haarnoja2018soft, lanier2025adaptingworldmodelslatentstate}. However, both model-free and model-based approaches remain challenging to apply reliably in low-data regimes and typically require sophisticated regularization. Efficient model-free methods aggressively reuse off-policy data and rely on frequent critic updates \cite{rlpd, droq, serl, redq}, often leading to value overestimation and unstable learning. Prior work mitigates these issues using conservative value estimation \cite{kumar2020conservativeqlearningofflinereinforcement}, policy constraints \cite{lei2025rl, zhang2025rewindlanguageguidedrewardsteach, smith2024grow}, or critic ensembles \cite{chen2021randomized, haarnoja2018learning, hiraoka2022dropout}. Model-based methods instead learn dynamics and reward models \cite{hafnerdream, hansen2022temporal} to reason about unseen trajectories, but must carefully avoid exploiting model inaccuracies. Common strategies include uncertainty-aware dynamics \cite{chua2018deep, levy2025learning, janner2021trustmodelmodelbasedpolicy} and explicitly penalizing out-of-distribution predictions \cite{yu2020mopo, yu2021combo, feng2023finetuningofflineworldmodels}. Rather than constraining learning, we show that bootstrapping a world model in simulation provides the coverage necessary to enable generalization beyond a small real-world dataset.

\textbf{Adaptation with Physics-based Models.} Many lines of work leverage approximate physics-based models for adaptation and control, but typically rely on simplified, low-dimensional state representations that are brittle in partially observed, contact-rich settings. Classical adaptive control \cite{aastrom1995adaptive, slotine1987adaptive, ioannou1996robust} and model-predictive control \cite{morari1999model, 8594448} approaches use highly simplified dynamics, abstracting away complex contacts and interactions. Neural physics engines \cite{xu2025neural, pfrommer2021contactnets} combine structured system identification with residual learning to adapt high-fidelity simulators using real-world data, but often assume access to object poses, contact labels, or reliable state estimates that degrade under partial observability. Closely related work learns world models from mixtures of simulated and real data \cite{li2025offline, xiao2025anycar} or transfers value functions from simulation to guide real-world learning \cite{westenbroek2022lyapunovdesignrobustefficient, yin2025rapidly}, but similarly depends on low-dimensional state observations. We instead propose distilling simulator structure from raw perception.

\textbf{Generative World Models for Robotics.} Recent work trains large video models on internet-scale data to learn broad physical priors for robotics \cite{yang2023learning, bruce2024genie, parkerholder2024genie2}, sometimes augmented with simulation data \cite{yang2023learning}. Translating these predictions into executable robot actions typically requires expert demonstrations, either by planning in video space and using inverse models to recover actions \cite{xie2025latent, jang2025dreamgen}, or by combining video prediction with behavior cloning \cite{zhu2025unified, li2025unified, cheang2024gr}. While effective for reproducing demonstrated behaviors, these methods are typically trained on narrow, expert-like action distributions and remain constrained by the real world data. In contrast, \Method\ does not predict pixels and learns a task-oriented latent world model with reward and value heads over a broad distribution of low-level robot actions generated in simulation, enabling planning to improve beyond the real-world data.

\section{ Preliminaries}

\textbf{Problem Setting.} We consider the problem of controlling a robotic system operating in the real world under partial observability and unmodeled dynamics. Specifically, we assume the dynamics are of the form $s_{t+1} \sim p(\cdot | s_t,a_t)$ where $s_t \in \mathcal{S}$ is the underlying state of the system and $a_t \in \mathcal{A}$ is the robot action. The underlying state is not directly observable in the real world as the robot only has access to raw observations $o_t \in \mathcal{O}$. We model the control problem as a partially observable Markov decision process (POMDP) defined by the tuple $(\mathcal{S}, \mathcal{A}, \mathcal{O}, p, r, \gamma)$, with user-specified reward function $r(s_t, a_t, s_{t+1})$ and discount factor $\gamma \in \left(0, 1\right]$, and seek to maximize the discounted return $\mathbb{E} \left[ \sum_{t=0}^{\infty} \gamma^t r(s_t, a_t, s_{t+1})\right]$. In general, dense informative reward functions are difficult to evaluate directly in the real world, given the difficulty in measuring the underlying state of the system. To render this problem more tractable, we make the following assumption: 
\begin{assumption}
We assume access to an approximate physics-based simulator $s_{t+1} \sim p_{\mathtt{sim}}(\cdot \mid s_t, a_t)$ that provides privileged access to the underlying state $s_t$. 
\end{assumption}

\textbf{Planning-Oriented Latent World Models.} We build on common planning-oriented latent world models from works such as \cite{hansen2022temporal, hafnerdream}. These approaches learn reward and value models that operate directly on raw observations, enabling planning in the real world using the algorithms outlined below. The primary novelty of \Method \ lies in our systematic framework for sim-to-real pre-training and adaptation. We structure our world model according to \cref{eq:model}, depicted in \cref{fig:model}:
\begin{equation}
\label{eq:model}
\begin{array}{ll}
    \text{Latent representation:} & \latent_{t} = \encoder_{\theta}(o_t)\\
    \text{History representation:} & h_t = \contextencoder_{\theta}(o_{t-H:t-1}, a_{t-H:t-1})\\
    \text{Latent dynamics:} & \hat{\latent}_{t+1:t+T} = \dynamicsmodel_{\theta}(\latent_{t}, \action_{t:t+T -1}, h_t)\\
    \text{Reward Prediction:} & \hat{\reward}_{t:t+T-1} = \rewardmodel_{\theta}(\hat{z}_{t:t+T},a_{t:t+T-1})\\
    \text{Value Prediction:} & \hat{\val}_{t+1:t+T} = \valuemodel_{\theta}(\hat{\latent}_{t:t+T})\\
    \text{Base Policy:} & \hat{\action}_{t:t+H} = \bcpolicy_{\theta}(\latent_{t}, h_t).
\end{array}
\end{equation}
Here $E_{\theta}$ encodes a latent representation $z_t$ of the state $s_t$ from raw observation $o_t$; $h_t$ is an encoding of history over a window of $H$ timesteps; $\hat{z}_{t+1:t+T}$ is a predicted sequence of $T$ future latent states generated by the learned model $f_\theta$; and $\hat{r}_{t:t+T-1}$ and $\hat{v}_{t+1:t+T}$ are reward and value estimates. We discuss key architecture decisions in \cref{sec:modeldetails}. 

\textbf{Sampling-Based Planning.}
We control the robot in the real world with \ac{mppi} \cite{williams2016aggressive} control, a sampling-based \ac{mpc} method. At each control step, we sample a batch of candidate future action sequences and evaluate them using the world model \cref{eq:model}, extracting the predicted future reward sequence $\hat{r}_{t:t+T-1}$ and terminal value $\hat{v}_{t+T}$, which are combined to compute the trajectory return \mbox{$\mathcal{R}(a_{t:t+T-1}) = \gamma^T\hat{v}_{t+T} + \sum_{s=t}^{t+T-1} \gamma^{s-t}\hat{r}_s$}.
MPPI then computes the control action to execute by importance-weighting sampled trajectories according to their predicted returns. Similar to TD-MPC \cite{hansen2022temporal}, we warm-start sampling by seeding a subset of candidate action sequences with noise-corrupted outputs of $\hat{a}_{t:t+T-1}$ from the base policy $\pi_\theta$.

\section{Simulation Distillation for Efficient Real-World Adaptation}

We now introduce \texttt{Simulation Distillation}, a framework for distilling physical priors and task structure into a world model in a form that enables rapid real-world adaptation with online planning and dynamics adaptation.

\subsection{Pretraining on Simulated Data} \label{sec:pretrain}
 
\Method\ builds on sim-to-real data-generation pipelines that use privileged, state-based expert policies to collect large-scale datasets paired with raw perception \cite{yin2026emergent}. While prior work primarily uses this data for imitation, planning-based adaptation places stronger demands on the learned model. A planner will actively search for high-value action sequences and exploit model errors wherever coverage is weak, so the model must remain reliable far beyond the expert and real-world data distributions. Dynamics predictions must generalize from few real-world samples, while reward and value models must distinguish good and bad outcomes under off-policy actions. To provide this coverage, we deliberately inject sub-optimal actions during simulation rollouts, providing coverage over mistakes, recoveries, and failed attempts.

\textbf{Expert Policy Training.} We first train a state-based expert policy $\pi^e(s_t)$ with reinforcement learning using existing sim-to-real pipelines  \cite{yin2026emergent, lee2020learning}. We also save the optimal state-based value function $V^e(s_t)$ and intermediate policy checkpoints $\{\pi^k\}_{k=1}^{K}$ learned during training to be used for value supervision and diverse, sub-optimal data generation.  

\textbf{Generating Diverse Trajectories and Dense Supervision.} We generate diverse simulation rollouts (\cref{fig:method}) by alternating between an expert policy $\pi^e$ and a set of sub-optimal policies $\{\pi^k\}_{k=1}^K$, and by periodically injecting random action perturbations over short temporal windows. This generates diverse failure and recovery behaviors beyond the nominal expert manifold, broadening coverage over dynamically feasible state-action trajectories and enabling the world model to reliably predict counterfactual outcomes. This results in a dataset $\simdataset = \{(o_t, a_t, r_t, v_t)\}_{t=0}^N$, where rewards $r_t$ are computed from privileged simulator state and value targets $v_t$ are provided by the expert value function $V^e(s_t)$. Crucially, this data-generation process is massively parallelizable and exploits the full training artifact of the simulator—including expert policies, intermediate checkpoints, and value functions—to produce rich supervision at scale. In addition, we randomize perceptual observations extensively to ensure robust transfer of the encoder. See \cref{app:datagendetails} for details.

\textbf{World Model Pretraining.}
We pretrain the world model \cref{eq:model} on $\mathcal{D}_{\text{sim}}$ by applying the following loss to predictions made at each time step $t$: 
\begin{equation}
\label{eq:simloss}
\begin{aligned}
\vspace{-.1em}
    \mathcal{L}_t^{\mathtt{sim}}(\theta) = \sum_{i=0}^{T}\bigg(\,\, &\underbrace{\norm{\hat{z}_{t+i+1} - \mathtt{sg}(E_{\theta}(o_{t+i+1}))}^2_2}_{\text{latent dynamics}}\\
    &+ c_1\underbrace{(\hat{r}_{t+i} - r_{t+i})^2}_{\text{reward}} + \, c_2\underbrace{(\hat{v}_{t+i+1} - v_{t+i+1})^2}_{\text{value}}\\
    &+ c_3 \underbrace{\mathds{1}_e(a_{t+i}) \norm{\hat{a}_{t+i} - a_{t+i}}^2_2}_{\text{behavior cloning}}\bigg),
\end{aligned}
\end{equation}
where $\mathtt{sg}$ is the \texttt{stop-grad} operator, constants $c_{1:3}$ are determined by normalizing over the range for each target, and $\mathds{1}_e(a_t)=1$ if $\action_t$ came from the uncorrupted expert policy and $\mathds{1}_e(a_t)=0$ otherwise. We apply various data augmentations discussed in \cref{app:manip,app:qped} to prevent overfitting to simulated observations an ensure that the model learns robust, transferable representations. In contrast to typical online MBRL approaches \cite{hansen2022temporal,hafnerdream} which are computationally intensive and attempt to bootstrap behavior from scratch, we offload behavior generation to a privileged expert. This reduces pretraining to optimizing a simple, stationary objective, without requiring temporal-difference learning. 

\begin{algorithm}[t]
\caption{\Method}
\label{alg:method}
\begin{algorithmic}[1]
\State \texttt{// Pretraining}
\State Perform RL, saving expert policy $\pi^e$, policy checkpoints $\{\pi^k\}_{k=1}^{K}$, and learned value function $V^e$
\State Generate dataset $\simdataset$ in simulation per \cref{sec:pretrain}
\While{not converged}
    \State Draw segments $\{(o_t, r_t, v_t, a_t)\}_{t=i-H}^{i+T} \sim \simdataset$
    \State Update $\theta$, minimizing $\mathcal{L}^{\mathtt{sim}}_t(\theta)$ at each time step
\EndWhile
\State \texttt{// Iterative Finetuning}
\State Initialize $\realdataset$ with offline real-world data if available
\For{$J$ iterations}
    \State \parbox[t]{\linewidth}
    {Collect real-world rollouts $\{(o_t, a_t)\}_{t=0}^M$ with MPPI \\
    and world model \cref{eq:model}, then add to $\realdataset$
    \vspace{3pt}}
    \While{not converged}
        \State Draw segments $\{(o_t, a_t)\}_{t=i-H}^{i+T} \sim \realdataset$
        \State Freeze $\contextencoder_\theta, \encoder_\theta, \rewardmodel_\theta, \valuemodel_\theta, \bcpolicy_\theta$
        \State Update $\dynamicsmodel_{\theta}$, minimizing $\mathcal{L}^{\mathtt{real}}_t(\theta)$
    \EndWhile
\EndFor
\end{algorithmic}
\end{algorithm}

\textbf{Robust Representations Without Reconstruction.}
Predicting rewards and values from $\latent$ forces the encoder to capture the task-specific features required for effective planning \cite{hansen2022temporal}. In contrast, many world-model approaches use pixel-level reconstruction objectives, arguing that they produce more robust and generalizable latent representations \cite{bruce2024genie, hafnerdream}. We find reconstruction unnecessary, and in some cases, harmful, for two reasons. First, our diverse data-generation procedure exposes the model to an extremely broad range of state-action pairs, spanning initial conditions, object configurations, contact modes, expert behaviors, failures, and recoveries. This diversity forces the model to learn a robust representation without the added computational cost of reconstructing pixels. Second, sim-to-real transfer requires extensive visual randomization. Pixel reconstruction would therefore pressure the latent state to encode randomized texture, lighting, and rendering artifacts that are deliberately varied and irrelevant to the task, which can hurt transfer~\cite{zhu2023repo}. 
\subsection{Real World Transfer and Efficient Dynamics Adaptation}
\label{sec:finetune}
Our key insight is that global task structure is largely invariant to low-level sim-to-real dynamics gaps. For example, in Peg Insertion (\cref{fig:front}), a meaningful latent state captures the locations of the peg and hole, while the value function encodes distance to the goal and motions leading to successful insertion. This structure persists across the sim-to-real gap, even though the low-level actions required to realize these behaviors differ between domains. Exploiting this decomposition, \Method\ finetunes only the environment-specific dynamics model while freezing the encoder, reward, and value functions. Planning with frozen reward and value models enables immediate improvement as dynamics predictions become more accurate, without requiring reward or value bootstrapping in the real world. Because the world model can be trained on arbitrary trajectories, \Method\ naturally supports off-policy learning and can incorporate prior data such as demonstrations. Importantly, the adaptation remains relatively short horizon and local, since no long-horizon bootstrapping is needed.

\textbf{Dynamics Adaptation Loss.} When updating the dynamics model we apply the following loss at each time $t$: 
\begin{equation}
\label{eq:realloss}
\begin{array}{l}
    \mathcal{L}_t^{\mathtt{real}}(\theta) = \sum_{i=0}^{T} \,\, \norm{\hat{z}_{t+i+1} - \mathtt{sg}(E_{\theta}(o_{t+i+1}))}^2_2, \\
    \text{with } C_\theta, E_\theta, R_\theta, V_\theta, \pi_\theta \ \text{frozen, } f_{\theta} \text{ finetunable}. 
\end{array}
\end{equation}
Because the encoder is frozen, it provides consistent latent targets 
$E_\theta(o_{t+i+1})$ throughout finetuning, avoiding the need to bootstrap a representation as in \eqref{eq:simloss}. This also anchors the adapted dynamics to the same latent representation used by the frozen reward and value models, rather than drifting away from the representation $R_\theta$ and $V_\theta$ were trained to evaluate.

\textbf{Iterative Improvement.} The overall pipeline for \Method \ is shown in \cref{fig:method} and in pseudo-code in \cref{alg:method}. \Method \ autonomously improves in the real-world by repeatedly collecting $M$ on-policy rollouts under the planner, adding this data to the real-world data set $\realdataset$, then finetuning $f_\theta$ to minimize prediction losses as in \eqref{eq:realloss}. Notably, because system identification can effectively learn from any real-world trajectories, \Method \ is a simple off-policy reinforcement learning strategy which can easily incorporate diverse data sources such as demonstrations or play data into $\realdataset$.

\begin{remark}
In contrast to standard MBRL frameworks \cite{hafnerdream,hansen2022temporal}, which must jointly bootstrap latent representations, value functions, and policies from scarce in-domain data, \Method\ offloads these challenging objectives to simulation, where diverse data is cheap and plentiful. As a result, we can reduce real world adaptation to supervised finetuning of the dynamics. 
\end{remark}

\subsection{World Model Design Decisions}
\label{sec:modeldetails}

In order to successfully improve behavior, the planner must sample numerous off-policy rollouts and accurately model returns. The following decisions were necessary to accelerate inference and enable reliable real-time decision making. 

\textbf{Minimal History Representation.}
Observation histories $o_{t-H:t}$ can contain high-dimensional inputs such as images that are costly to process. To reduce inference cost, we split observations $o_{t} = (o_t^p,o_t^e)$ into proprioceptive and exteroceptive components and feed only $(o_{t-H:t}^p,a_{t-H:t},o_t^e)$ into the history encoder $C_\theta$. Using only the most recent high-dimensional observation substantially reduces planning latency and, empirically, improves training stability by reducing context length.

\textbf{Chunked Prediction for Planning.} Autoregressive world models \cite{hansen2022temporal, hafnerdream, chen2022transdreamer} require sequential unrolling over the planning horizon, which bottlenecks parallelism when  evaluating numerous rollouts. Inspired by \cite{xiao2025anycar}, our latent dynamics model $f_\theta$ predicts $T$ future states in a single forward pass using a transformer with cross-attention between the encoded history tokens and a candidate action sequence, together with a causal mask. This chunked prediction fully exploits GPU parallelism, dramatically improving planning throughput. 

\textbf{Sequence-to-Sequence Return Modeling.} Prior work often decodes rewards and values with per-timestep MLPs applied independently to each predicted latent state \cite{hansen2022temporal, hafnerdream, chen2022transdreamer}. We instead use transformer-based reward and value models that attend over the entire predicted latent trajectory $\hat{z}_{t:t+T}$. This aggregates information across the entire trajectory, yielding more accurate return estimates (see \cref{sec:ablations}).

\section{Experiments}
\label{sec:exp}

\begin{figure*}
    \centering
    \includegraphics[width=0.99\textwidth]{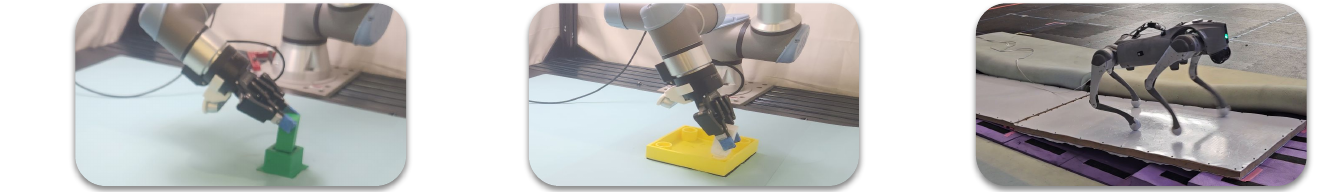}
    \includegraphics[width=0.99\textwidth]{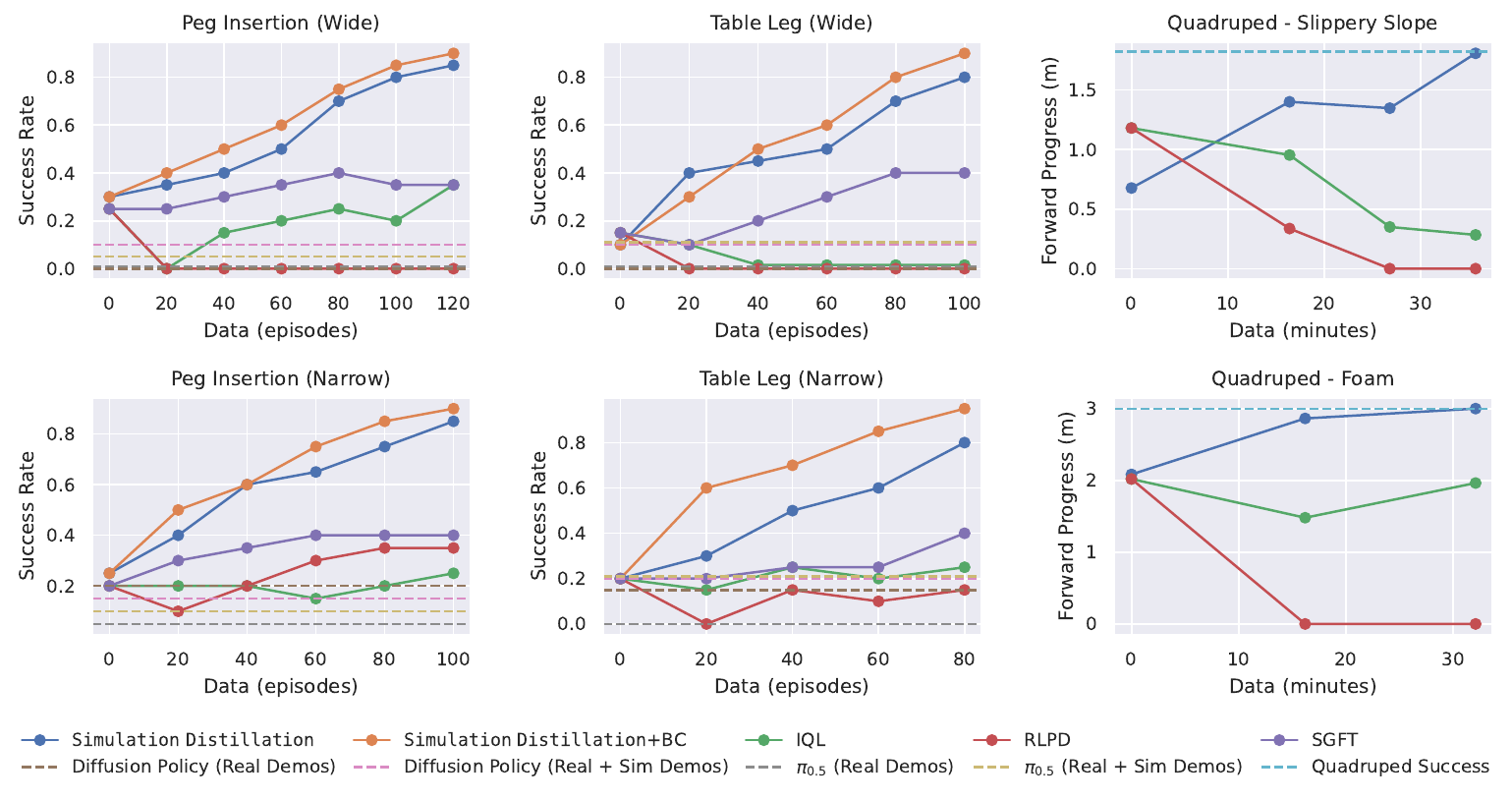}
    \vspace{-1em}
    \caption{\footnotesize{
    Real-world results. Success rate for two manipulation tasks, computed over 20 trials, and average forward progress for two quadruped locomotion tasks, averaged across all $15$ trials (3 speeds, 5 trials each), as a function of real-world finetuning data. \Method\ exhibits rapid and consistent improvement with limited data by finetuning only the latent dynamics model while planning with frozen reward and value models. In contrast, direct policy finetuning with the baselines shows limited or no improvement under the same data budgets.
    }}
    \vspace{-1em}
    \label{fig:results}
\end{figure*}

We evaluate the ability of \Method\ to adapt in the real world on the four manipulation and quadruped tasks depicted in \cref{fig:front} and carefully ablate key design decisions. Additional details can be found in the Appendix. We aim to answer: $1)$ does \Method \ outperform existing online RL methods and behavior cloning baselines? $2)$ what factors enable the planner to effectively improve performance with minimal real-world data?  $3)$ which components of the architecture and pretraining procedure are crucial for the performance of \Method?  

\subsection{Robotic Systems and Tasks}
\textbf{Manipulation System and Tasks. } Manipulation experiments are conducted on a UR5e robot. Actions are six-dimensional relative end-effector pose targets and a binary gripper action, and observations include joint states and three $224\times224$ RGB images from wrist-mounted, overhead, and side-view cameras. Each image is encoded with a pretrained ResNet-18, fused with proprioception, and mapped to a 64-dimensional latent state $z$. Training uses 100k trajectories (see \cref{app:manip} for details on data mixture). We use history and prediction horizons $H=T=5$ and control the robot at \SI{5}{\hertz}. Expert policies are trained with \cite{yin2026emergent}. We consider two precise assembly tasks, with initial conditions drawn from a Narrow (\qtyproduct{2 x 2}{\cm}) or Wide (\qtyproduct{35 x 35}{\cm}) grid:
\begin{hangingpar}
\textbf{1) Peg Insertion.} We construct a peg insertion task similar to the \SI{16}{\mm} square peg task from \cite{narang2022factory}. This task requires picking the peg, aligning it with the hole, and insertion.
\end{hangingpar}
\begin{hangingpar}
\textbf{2) Table Leg.} We follow \cite{heo2023furniturebench}, wherein a table leg must be picked, aligned, and threaded into a hole on a table.
\end{hangingpar}
\textbf{Quadruped System and Tasks.} We conduct quadrupedal experiments on a Unitree Go2, with actions given as position targets for the $12$ joints. Observations include proprioception and a local terrain height map, encoded using a CNN and fused with low-dimensional observations via an MLP to produce the latent state $z$. The policy $\pi_\theta$, reward $R_\theta$, and value $V_\theta$ are additionally conditioned on desired base forward, lateral, and yaw velocities. Pretraining uses 100M simulated trajectories. We use history and prediction horizons $H=T=25$ and plan at \SI{50}{\hertz} on a laptop with an RTX 4090M. See \cref{app:qped} for details. We consider two tasks: 
\begin{hangingpar}
\textbf{1) Slippery Slope.} The robot traverses straight over two panels inclined at $3.0^\circ$ and $5.7^\circ$, respectively. The panels are covered with PTFE (Teflon), and the robot’s feet are wrapped with thermoplastic, creating extremely low-friction contacts. A trial is successful if the robot traverses \SI{1.82}{\meter}, clearing both panels. We conduct five trials per commanded forward speed at $0.1$, $0.3$, and $0.5$~\si{\meter\per\second}.
\end{hangingpar}

\begin{hangingpar}
\textbf{2) Foam.} The robot traverses two \SI{5}{\cm} thick overlapping memory foam pads whose compliant dynamics are not modeled in simulation. A trial is  successful if the robot traverses \SI{3.00}{\meter} to clear the foam. We conduct five trials per commanded forward speed at $0.2$, $0.7$, and $1.2$~\si{\meter\per\second}.
\end{hangingpar}

\subsection{Real-World Learning Methods.}

We evaluate \Method\ against state-of-the-art real-world RL methods (and behavior cloning baselines on manipulation tasks). All RL methods use the same encoders as \Method, but with an MLP policy head (no action chunking). 

\textbf{Manipulation.}
For each task, we collect $20$ real-world demonstrations via teleoperation. We evaluate two variants of \Method: (i) \Method, which adapts only the latent dynamics model using on-policy rollouts (\cref{alg:method}), without access to the demonstrations and (ii) \Method+BC, which additionally finetunes the base policy $\pi_\theta$ using expert action labels from the teleoperated data. In both cases, the dynamics model is updated after every 20 real-world episodes. We compare against model-free RL baselines for offline-to-online RL, RLPD \cite{rlpd} and IQL \cite{kostrikovoffline}, trained from sparse rewards to reflect common manipulation settings where dense rewards are difficult to specify; these baselines update after every episode. We also include SGFT-SAC \cite{yin2025rapidly}, a model-free method that transfers a simulated value function, isolating the benefit of full world-model adaptation beyond value transfer alone. All online RL methods are given access to the 20 pre-collected demonstrations. Finally, we compare against behavior cloning baselines—Diffusion Policy \cite{chi2024diffusionpolicy} and $\pi_{0.5}$ \cite{intelligence2025pi05}—trained with 100 real-world demonstrations, as well as variants trained on both real-world demonstrations only and co-trained on these demonstrations and the simulated dataset used by \Method.  The task set-up is identical to the corresponding tasks from \cite{yin2025rapidly}, however, we depart from this work by defining a rollout to be successful if it completes the task within $45$ seconds. 

\textbf{Quadruped.}
For quadruped locomotion, we compare \Method\ against the off-policy algorithm RLPD \cite{rlpd} and the offline-to-online finetuning method from IQL \cite{kostrikovoffline}, whose value function is pretrained in simulation. All methods use the same learned reward model from \Method, allowing us to isolate the effect of different adaptation strategies. We do not have the ability to collect demonstrations for this system, and thus do not compare methods which require them. 

\subsection{Real World Improvement Results}

\Cref{fig:results} summarizes our results. Across all tasks, \Method\ consistently outperforms prior approaches, achieving substantially higher success rates with far greater sample efficiency than online RL baselines, while autonomously improving well beyond the performance of behavior cloning methods. Across the board \Method \ typically reaches scores around $2\times$ higher than any baseline. Standard RL finetuning methods frequently exhibit catastrophic forgetting, with performance collapsing during adaptation, whereas \Method\ makes steady, monotonic progress throughout training, due to its ability to side step long-horizon credit assignment and improve performance with simple supervised learning. SGFT avoids catastrophic collapse by transferring value functions from simulation, but remains significantly more sample inefficient than \Method, which can efficiently improve by leveraging the world model to make numerous counterfactual predictions about trajectories the robot has not directly experienced. Finally, we observe that providing \Method \ with demonstrations only boosts performance, highlighting how \Method \ can naturally absorb heterogeneous, mixed quality sources of data.

\begin{figure}
    \centering
    \includegraphics[width=0.48\textwidth]{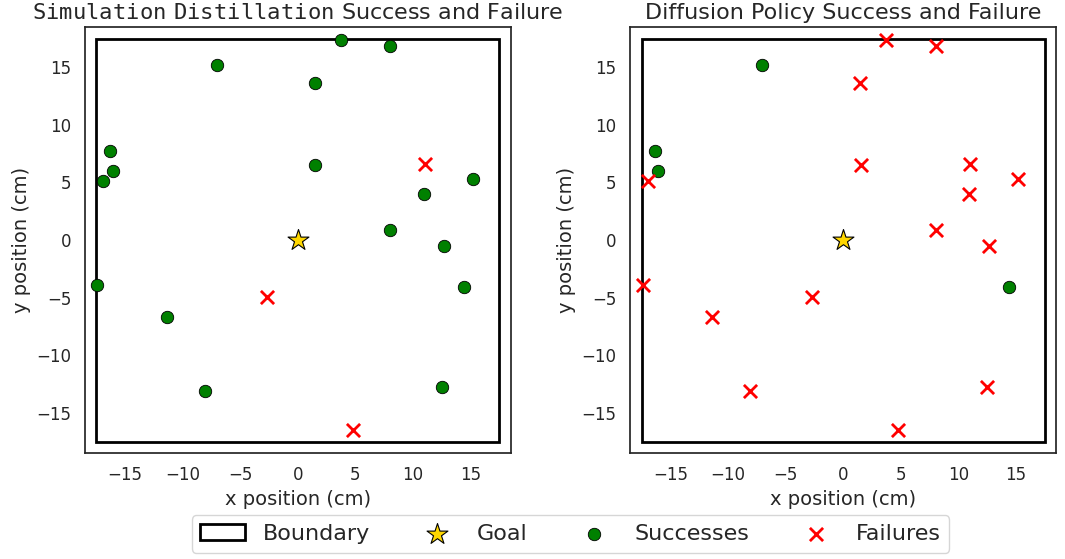}
    \vspace{-0.5em}
    \caption{
    \footnotesize{Scatter-Plot showing successful and failed attempts at solving the Peg Wide task for Diffusion Policy (Right) and the final trained policy for \Method (Left). The broad coverage of pretraining data for \Method \ enables efficiently learning policies which are far more robust than baselines.}
    }
    \label{fig:scatter}
    \vspace{-1em}
\end{figure}

For the two manipulation tasks, the performance gap between \Method\ and baselines widens as the task is made more difficult by expanding the ranges of initial conditions from the narrow to wide distribution. This underscores the benefit of broad simulation pretraining, which enables \Method\ to retain structural priors from the simulator and reliably improve performance over the entire state-space with limited real-world data. This effect is illustrated in \cref{fig:scatter}, which visualizes successful and failed initial conditions for \Method\ and Diffusion Policy on Peg Hard, revealing the substantially greater robustness of policies learned by \Method.

\begin{figure}
    \centering
    \includegraphics[width=0.48\textwidth]{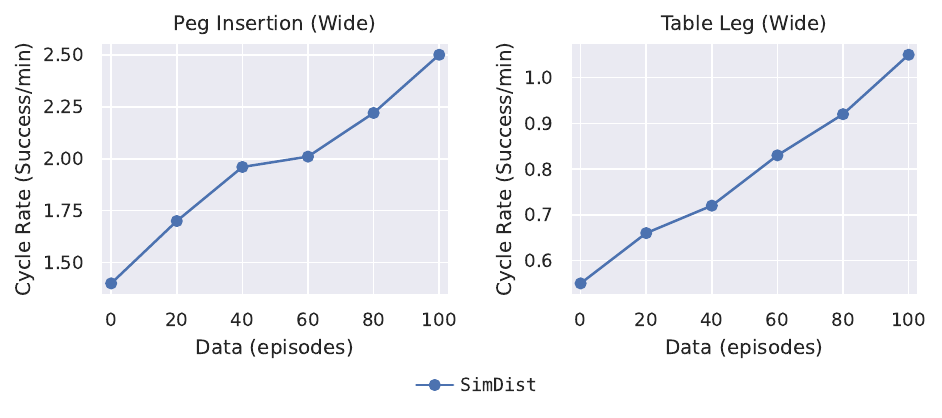}
    \vspace{-0.5em}
    \caption{
    \footnotesize{Throughput of manipulation policies throughout training. \Method \ reliably improves the velocity of successful task completions. 
    }
    }
    \label{fig:cycle}
    \vspace{-1em}
\end{figure}

Finally, \cref{fig:cycle} provides additional insight into how \Method \ improves performance by plotting the number of success/min achieved during training. \Method \ monotonically improves throughput by $\sim\!1.5{\times}$--$2{\times}$ over zero-shot performance. 

\begin{figure}
    \centering
    \includegraphics[width=0.48\textwidth]{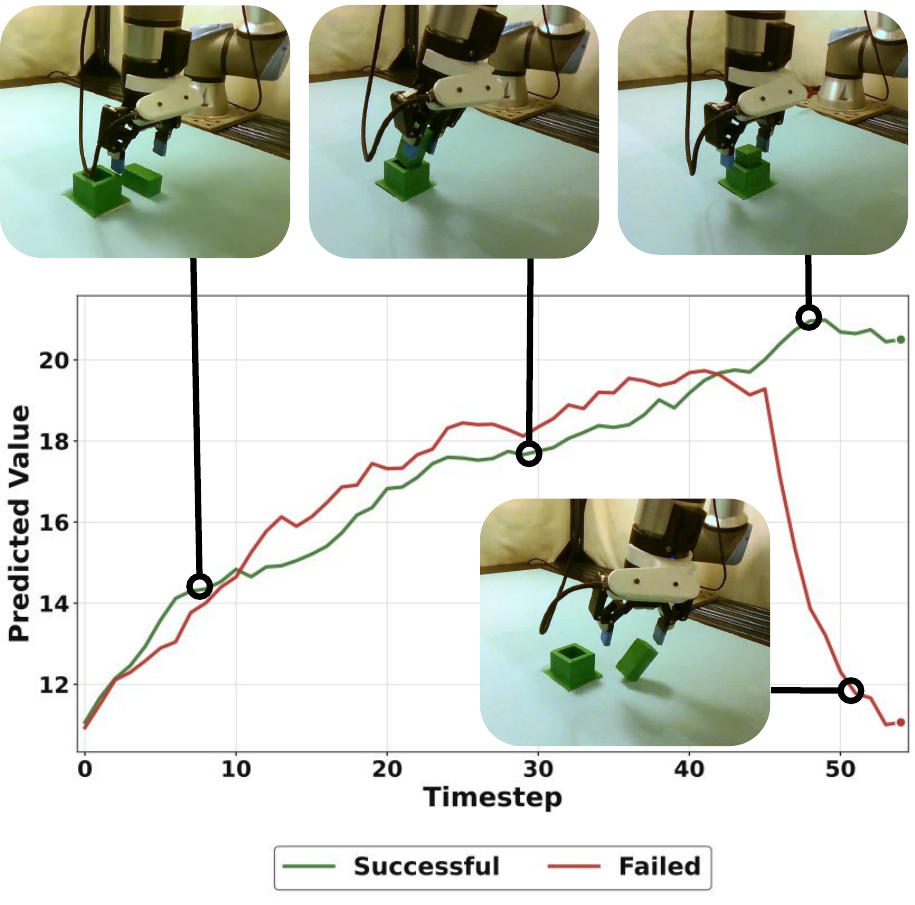}
    \vspace{-0.5em}
    \caption{\footnotesize{Value predictions from \Method\ along successful and failed real-world Peg trajectories starting from the same initial condition. The predicted values track task progress and clearly distinguish successful from failure.}}
    \label{fig:value}
\end{figure}

\begin{figure*}
     \centering
     \begin{subfigure}[t]{0.30\textwidth}
         \centering
         \panel{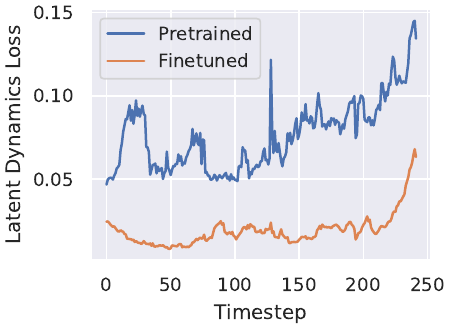}{(a)}{\linewidth}
         \phantomsubcaption
         \label{fig:consistloss}
     \end{subfigure}
     \hfill
     \begin{subfigure}[t]{0.140\textwidth}
         \centering
         \panel{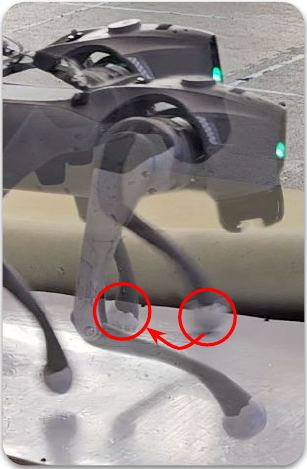}{(b)}{\linewidth}
         \phantomsubcaption
         \label{fig:footslip}
     \end{subfigure}
     \hfill
     \begin{subfigure}[t]{0.235\textwidth}
         \centering
         \panel{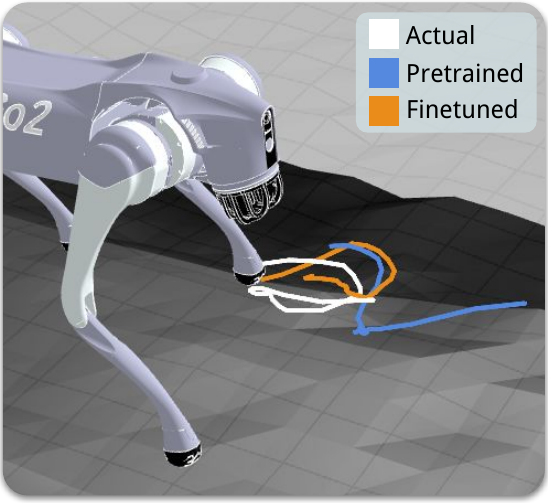}{(c)}{\linewidth}
         \phantomsubcaption
         \label{fig:footpreds}
     \end{subfigure}
     \hfill
     \begin{subfigure}[t]{0.295\textwidth}
         \centering
         \panel{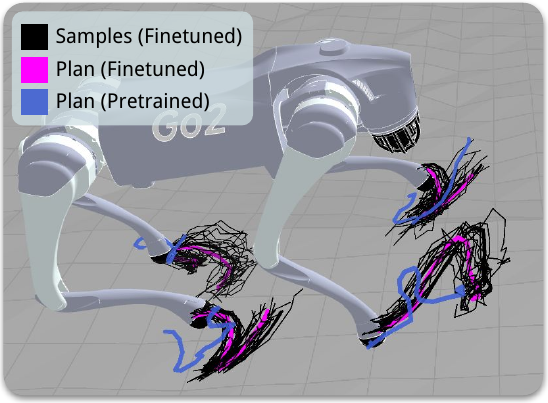}{(d)}{\linewidth}
         \phantomsubcaption
         \label{fig:planning}
     \end{subfigure}
     \hfill
     \vspace{-1em}
     \caption{\footnotesize{
     \textbf{(a)} Finetuning drastically lowers dynamics prediction loss on a quadruped Slippery Slope trial.
    \textbf{(b)} Frames showing the front left foot slipping during the trial.
    \textbf{(c)} Foot-trajectory predictions from the world model at the same instant: the finetuned model correctly anticipates future slippage, while the pretrained model fails to do so.
    \textbf{(d)} Visualization of sampling-based planning. Candidate action sequences are evaluated with the world model. Sampled trajectories are shown from the finetuned dynamics model, along with the resulting optimal plans under the finetuned and pretrained models. The finetuned model produces plans that account for real world dynamics mismatch, while the pretrained model generates qualitatively different plans.
    }}
    \vspace{-1em}
\end{figure*}

\subsection{Analyzing Real-World Predictions and Effects on Planning}
\label{sec:ablations} The planner can only improve behavior if it can reliably distinguish trajectories with high and low returns. This requires both accurate dynamics predictions and successful transfer of reward and value models. We examine value transfer in \cref{fig:value}, which plots predicted values over time for successful and failed trajectories. For the successful rollout, predicted value increases consistently over time, while the  value drops sharply when the robot drops the peg during the failed rollout.  Thus the transferred encoder $E_\theta$ and value function $V_\theta$ \emph{reliably discriminate between successful and failed trajectories}.

\begin{figure}
    \centering
    \includegraphics[width=0.48\textwidth]{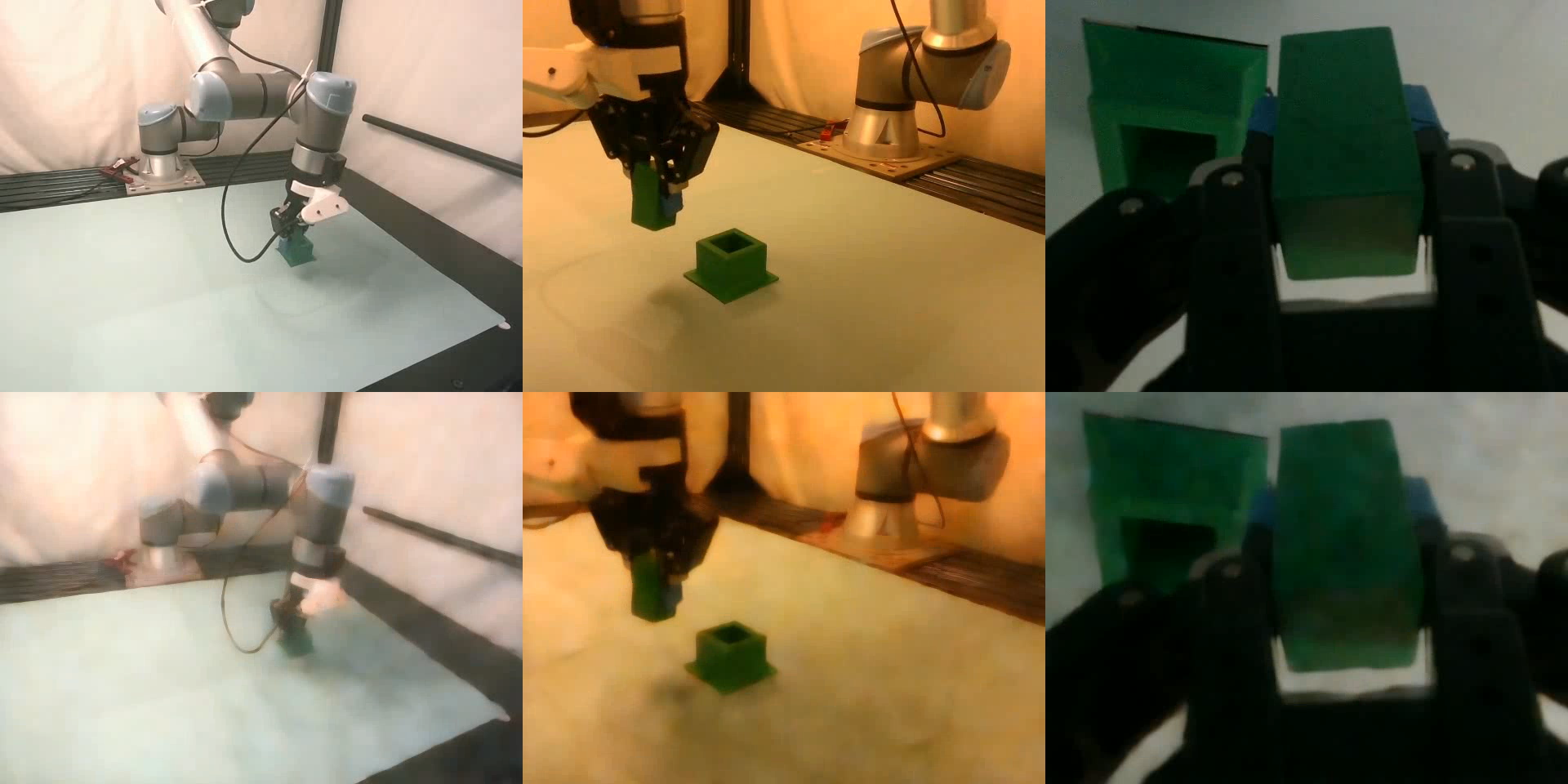}
    \vspace{-0.5em}
    \caption{
    \footnotesize{Real world camera observations and images reconstructed from the corresponding encoded latent state $z_t$ using an auxiliary encoder. We reiterate pixel reconstruction is not used as a training objective for the model. 
    }
    }
    \label{fig:reconstruction}
    \vspace{-1em}
\end{figure}

Next in \cref{fig:reconstruction} we compare ground truth camera observations for the peg task compared to images reconstructed from the corresponding latent encoding $z_t = E_\theta(o_t)$ generated by the frozen encoder. The images are generated by an auxiliary probe trained on all available real world data, and we reiterate that the world model itself is not trained with the reconstruction loss. Even though the encoder is not trained to explicitly reconstruct pixels, the the broad diverse pre-training of \Method \ forces the encoder to learn a robust representation which is able to capture the underlying state of the real-world scene, enabling the reconstructions portrayed.

We next evaluate dynamics prediction accuracy on the Quadruped Slippery Slope task in \cref{fig:consistloss}. We roll a held-out real-world trajectory through the world model and compute the latent dynamics loss \cref{eq:realloss} at each timestep, yielding an average loss of $0.076$ for the pretrained model and $0.019$ after finetuning. To visualize the impact of this improvement, we decode predicted latent states into predicted front-left foot positions in \cref{fig:footpreds}. Before finetuning, the model incorrectly predicts stable contact on the PTFE surface and fails to anticipate slip; after finetuning, it accurately predicts future slippage, closely matching the real trajectory (\cref{fig:footslip}). This illustrates how broad simulation pretraining enables the dynamics model to generalize to real-world trajectories outside the training set. 

Finally, improved dynamics predictions directly reshape planning behavior. As shown in \cref{fig:planning}, trajectory samples generated with the finetuned model reflect the altered contact dynamics and lead the planner to select plans that account for real-world slip. In contrast, plans derived from the pretrained model are qualitatively inconsistent with the true dynamics. Together, these results show that finetuning corrects latent dynamics errors in a way that meaningfully changes the planner’s trajectory distribution, explaining the rapid real-world performance gains observed in \cref{fig:results}.

\subsection{Ablating Key Design Decisions}

\begin{figure}
    \centering
    \includegraphics[width=0.48\textwidth]{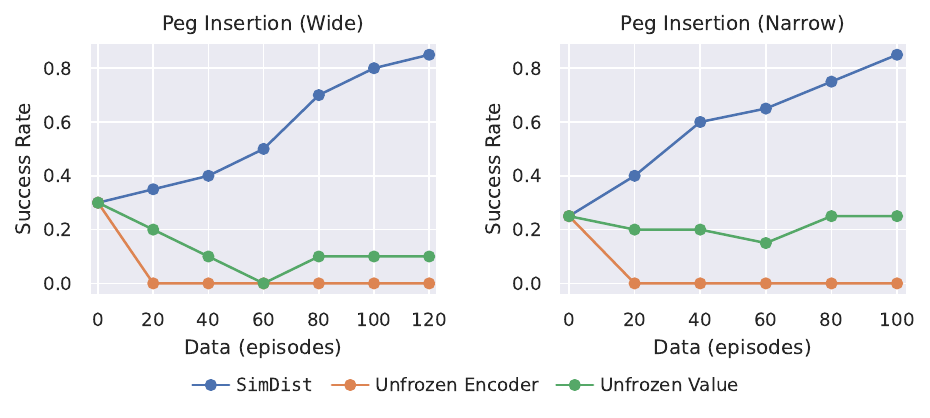}
    \vspace{-0.5em}
    \caption{
    \footnotesize{Unfreezing world-model components during real-world adaptation.}
    }
    \label{fig:unfreeze}
    \vspace{-1em}
\end{figure}

\textbf{Unfreezing World Model Components.} We first ablate unfreezing world-model components during real-world adaptation, with results in \cref{fig:unfreeze}. Unfreezing the encoder causes complete performance loss, as frozen reward and value heads receive latents outside their training distribution. Unfreezing the value function, following~\cite{feng2023finetuningofflineworldmodels}, reintroduces long-horizon credit assignment from limited real-world data and causes catastrophic forgetting. We omit reward-model unfreezing, since accurate dense reward labels are generally unavailable in the real world, underscoring a key advantage of \Method: reward transfer without real-world labels.

We run the following ablations in simulation, with results in \cref{tab:ablations} (see \cref{app:ablations} for details). 

\textbf{Data Scale and Diversity.}
We first study the effect of dataset scale by reducing simulation rollouts to $10\%$ and $50\%$ of the full pretraining data. Performance drops sharply as data is reduced for both systems. We next evaluate the role of data diversity by comparing pretraining on expert-only trajectories to our mixed datasets with equal data volume, and observe that expert-only data leads to a substantial performance drop. Together, these results highlight the importance of obtaining large, diverse datasets to provide the coverage needed to pretrain world models which can be used for effective planning. 

\begin{table}[htb]
    \centering
    \caption{
    \footnotesize{
    Ablation results in simulation, reporting success rates for manipulation tasks and average state-based reward per episode for the quadruped.
    }
    }
    \vspace{-1em}
    \begin{tabular}{lccc}
        ~ & Peg Insertion & Table Leg & Quadruped \\
        ~ & (SR) & (SR) & (Reward) \\ 
        \hline
        \Method & 0.90 & 0.85 & 22.78\\ 
        \hline
        50\% data & 0.72 & 0.61 & 22.73 \\ 
       10\% data & 0.06 & 0.02 & 19.38 \\ 
        Expert Data Only & 0.10 & 0.05 & 16.68 \\ 
        MLP Reward+Value Models & 0.82 & 0.60 & 19.47 \\ 
        Raw Obs. Reconstruction & 0.32 & 0.21 & 23.34 \\
        \hline
    \end{tabular}
    \label{tab:ablations}
    \vspace{-1em}
\end{table}

\textbf{Reward and Value Transformers.}
Next, we ablate the use of sequence-to-sequence transformers for reward and value prediction, replacing them with per-timestep MLP decoders as in \cite{hafnerdream, hansen2022temporal}. This consistently degrades performance across tasks. We attribute this to the inability of per-step models to capture trajectory-level structure, which is essential for accurately evaluating candidate action sequences during planning.

\textbf{Reconstruction Loss.} We study the effect of adding an observation reconstruction loss to the training objective, a common design choice in MBRL \cite{hafnerdream, bruce2024genie}. While reconstruction slightly improves quadruped performance, it leads to a steep performance drop for manipulation, consistent with the concern that pixel reconstruction pressures the latent state to encode task-irrelevant details (\cref{sec:pretrain}).

\section{Conclusion}
We introduced \Method, a framework for sim-to-real adaptation that uses the modular structure of world models to target the dynamics gap between simulation and reality. Across two precise manipulation tasks and two quadruped locomotion tasks, \Method\ improves more efficiently and reliably than prior RL finetuning methods, which often collapse or fail to make meaningful progress. However, freezing reward and value models can cap performance when the transferred value function saturates or no longer distinguishes high-performing real-world trajectories. Closing the gap to near-perfect success may require selectively updating value functions in addition to dynamics. \Method\ also does not train on internet-scale video or richer sensing modalities, and still depends on simulation coverage broad enough to support reliable planning.

\section*{Acknowledgments}
The authors thank Trey Smith, Brian Coltin and the members of the WEIRD Lab at UW for their insights and feedback. This work was supported by a NASA Space Technology Graduate Research Opportunity under award 80NSSC23K1192 and the generous support of FieldAI. 

\bibliographystyle{plainnat}
\bibliography{references}

 \newpage
\appendix
\crefalias{subsection}{appendix}

\subsection{Diverse Data Generation Details}
\label{app:datagendetails}
\Cref{alg:datagen} details the data generation process. Data generation proceeds with running many environments in parallel. For the $j$-th environment, we sample a diagonal action-noise covariance $\Sigma_j$, where each element on the diagonal is sampled between a minimum and maximum variance: $\sigma_i \sim \mathcal{U}[\sigma^\mathtt{min}_i, \sigma^\mathtt{max}_i]$. When each environment is reset, we sample contiguous noise intervals during which Gaussian noise is added to policy actions. In addition, each environment is assigned a randomly selected policy checkpoint from $\{\pi^k\}_{k=1}^{K}$, or the expert policy $\pi^e$. Together, policy mixing and temporally structured action noise produce a dataset that captures both expert-like trajectories and systematic deviations beyond the optimal manifold, which is critical for  world-model training. During rollouts, we query the optimal state-based value function $V^e$ at each visited state to generate value targets $v_t$ for distilling an approximate optimal value function (\cref{sec:pretrain}). We also record an expert action flag $\expertflag_t$ that distinguishes the expert actions from the expert policy $\pi^e$ versus noised or earlier-checkpoint actions to support behavior cloning (\cref{sec:pretrain}).

\subsection{Manipulation Experiment Details}
\label{app:manip}
\textbf{Expert Policy Training.} For our manipulation experts we use the expert policies $\pi^e$ and value function $V^e$ from \cite{yin2026emergent}, replicating the training from this work exactly. For the sake of brevity, we refer the reader to \cite{yin2026emergent} for more details, including the exact rewards we use.

\textbf{Data Generation.} To generate the simulation dataset $\simdataset$, environments are reset with sub-optimal policies $\{\pi^k\}_{k=1}^K$ with probability $0.5$, and Gaussian action perturbations are injected in contiguous intervals sampled from $\mathcal{U}[1, 5]$ steps, interleaved with noise-free intervals sampled from $\mathcal{U}[5, 10]$ steps. We generate 100k trajectories for each tasks, of which approximately 36\% optimal actions. We save policies every 100 checkpoints up to checkpoint 1000.

\textbf{World Model Structure.} The encoder for the world model first passes each of the three camera images through a ResNet-18 encoder pretrained on imagenet, producing embeddings of size $3 
\times 512$, which are stacked and concatenated with the robots 6 joint observations and then passed through an MLP to produce the latent $z$. Specific parameters are in  \cref{tab:model_man}.

\begin{table}[htb]
\caption{\footnotesize{World model architectural parameters for manipulation.}}
\centering
\begin{tabular}{lc}
\hline
Parameter & Value \\
\hline
Embedding dimension & 64 \\
All transformers MLP hidden size & 256 \\
Dynamics transformer layers & 3 \\
Dynamics transformer heads & 4 \\
Reward transformer layers & 1 \\
Reward transformer heads & 1 \\
Value transformer layers & 1 \\
Value transformer heads & 1 \\
Base policy transformer layers & 4 \\
Base policy transformer heads & 8 \\
\hline
\end{tabular}
\vspace{0.05cm}
\label{tab:model_man}
\end{table}

\begin{algorithm}[t]
\caption{Diverse Data Generation}
\label{alg:datagen}
\begin{algorithmic}[1]
\State \textbf{Input:} Expert policy $\pi^e$, policy checkpoints $\{\pi^k\}_{k=1}^{K}$, optimal value function $V^e$
\State \textbf{Output:} Simulation dataset $\simdataset$
\For{environment $j \in \{1\ldots N_E\}$}
    \State
    \parbox[t]{\linewidth}{Sample action noise variance: $\Sigma_j = \text{diag}(\sigma)$ \\ with $\sigma_i \sim \mathcal{U}[\sigma_i^\mathtt{min}, \sigma_i^\mathtt{max}]; i \in \{1,\ldots,\text{dim}(\mathcal{A})\}$ 
    }
\EndFor
\For{$t = 0$ to number of steps $N_T$}
\For{environment $j \in \{1\ldots N_E\}$}
    \If{$\simstate_t$ is terminal}
        \State Reset environment
        \State Sample noise intervals $\mathcal{T}_\mathtt{noise}$
        \State
        \parbox[t]{\linewidth}{Randomly select a policy $\pi^k$ from $\{\pi^k\}_{k=1}^{K}$ \\ and assign it to the environment.}

    \EndIf
    \State $\eps_t \sim \mathcal{N}(0, \Sigma_j)$ if $t \in \mathcal{T}_\mathtt{noise}$, else $\eps_t \gets 0$
    \State $\action_t \gets \pi^k(\simstate_t) + \eps_t$
    \State \parbox[t]{\linewidth}{$\expertflag_t \gets 1$ if (using expert $\pi^e$ and $\eps_t = 0$), \\ else $\expertflag_t \gets 0$}
    \State $\val_t \gets V^e(\simstate_t)$
    \State $(\simstate_{t+1}, \proprio_t, \reward_t) = \mathtt{EnvStep}(\simstate_t, \action_t)$
    \State Add $(\proprio_t, \action_t, \expertflag_t, \reward_t, \val_t)$ to $\simdataset$
\EndFor
\EndFor
\end{algorithmic}
\end{algorithm}

\textbf{World Model Pretraining.}
We pretrain the world model for two epochs over the full simulation dataset $\simdataset$. Using a batch size of 256 and approximately 200k gradient updates. Optimization is performed with Adam using an initial learning rate of \num{2e-4}, which is annealed to \num{1e-4} via a cosine decay schedule, with a linear warmup over the first 10{,}000 steps. We apply data augmentation by injecting zero-mean Gaussian noise into the proprioceptive observations, and visual augmentations such as color jitter,  gaussian blurring and random cropping.

\textbf{Hardware Deployment.} We use MPPI as implemented in TD-MPC \cite{hansen2022temporal} with the hyperparameters listed in \cref{tab:mppi_manip}.

\begin{table}[htb]
\caption{\footnotesize{MPPI parameters for manipulation.}}
\centering
\begin{tabular}{lc}
\hline
Parameter & Value \\
\hline
Candidate actions batch size & 250 \\
Noised base policy actions batch size & 100 \\
Solver iterations & 3\\
Initial action standard deviation & 1.0 \\
Minimum action standard deviation & 0.05 \\
Base policy action standard deviation & 0.1 \\
Elites & 64 \\
Temperature & 0.4 \\
Momentum & 0.0 \\
Discount & 0.99 \\
\hline
\end{tabular}
\vspace{0.05cm}
\label{tab:mppi_manip}
\end{table}

\subsection{Quadruped Experiment Details}
\label{app:qped}

\textbf{Expert Policy Training.} We train a state-based expert policy $\pi^e$ and its associated optimal value function $V^e$ using PPO in IsaacLab \cite{mittal2025isaaclab}. Both the policy and value networks are MLPs with three hidden layers of width 512 and operate on privileged simulator state variables listed in \cref{tab:state_qped}. The expert is trained with a dense state-based reward composed of the terms summarized in \cref{tab:reward_qped}; full implementation details will be released with the public code. To improve robustness and coverage, we apply domain randomization over the parameters in \cref{tab:state_qped} and randomize terrain conditions across steps, boxes, rough terrain, and slopes, using a curriculum that gradually increases terrain difficulty. Training is performed with 4096 parallel simulation environments over 5000 PPO iterations, with 24 environment steps per iteration, for a total of 490M environment steps. We save policy checkpoints at iterations $\{0, 50, 100, 150, 200, 250, 300, 400, 500, 1000, 2000\}$.

\textbf{Data Generation.} To generate the simulation dataset $\simdataset$, environments are reset with sub-optimal policies $\{\pi^k\}_{k=1}^K$ with probability $0.5$, and Gaussian action perturbations are injected in contiguous intervals sampled from $\mathcal{U}[1, 50]$ steps, interleaved with noise-free intervals sampled from $\mathcal{U}[25, 500]$ steps. We run 4096 parallel environments for 25000 steps, yielding approximately 100M data points, of which $55.7\%$ correspond to uncorrupted expert actions. Data generation takes approximately 7 hours with a single NVIDIA RTX 4500 Ada GPU.

\begin{figure*}[t]
    \centering
    \includegraphics[width=.99\textwidth]{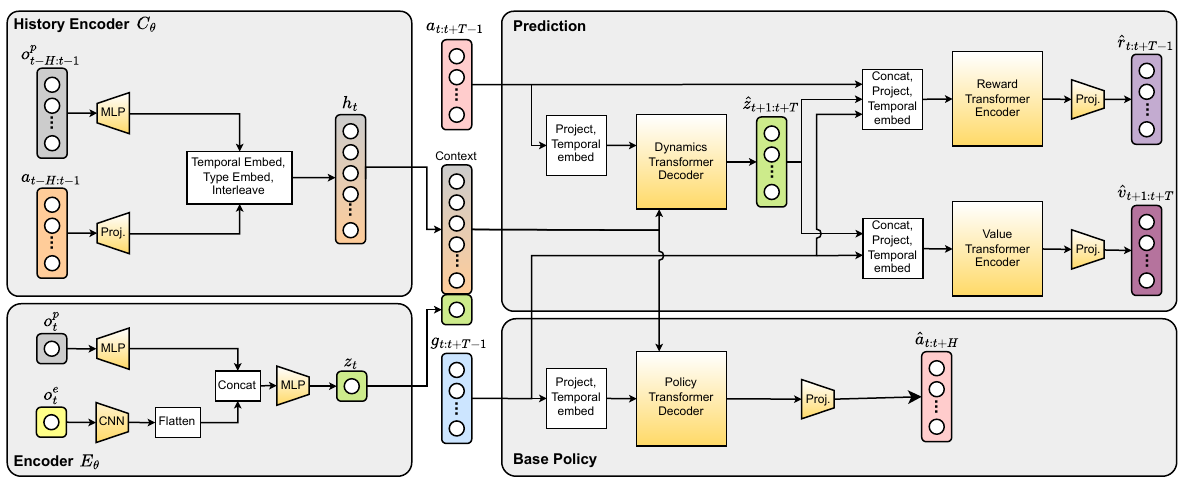}
    \caption{
    \footnotesize{Detailed world model architecture for the quadruped. 
    }
    }
     \vspace{-1em}
    \label{fig:model_qped}
\end{figure*}

\textbf{World Model Structure.} \Cref{fig:model_qped} illustrates the world model architecture, and \cref{tab:model_qped} lists the corresponding model parameters used in the quadruped experiments. The observation space for the quadruped is given in \cref{tab:obs_qped}. The history encoder processes a history of proprioceptive observations (all observations except the height map) and actions by first projecting each input, assigning a type embedding to distinguish observations from actions, and interleaving the resulting embeddings to form the history representation $h_t$. The latent encoder $E_\theta$ encodes the local terrain height map using a CNN followed by spatial encoding, flattening, and projection; this representation is concatenated with the projected latest proprioceptive observation and passed through an MLP to produce the latent state embedding. Commands consist of the desired forward, lateral, and yaw velocities of the base \mbox{$g_t := (v^x_t, v^y_t, \omega_t)$}.  The future commands $g_{t:t+T-1}$ are concatenated to the inputs of the base policy $\pi_\theta$, reward model $R_\theta$, and value model $V_\theta$.

\begin{table}[htb]
\centering
\caption{\footnotesize{Privileged simulator state space for the quadruped, along with domain randomization ranges, where applicable.}}
\begin{tabular}{lcc}
\hline
State Variable & Dim. & Domain Rand. \\
\hline
Base linear velocity & $3$ & - \\
Base angular velocity & $3$ & - \\
Projected gravity vector & $3$ & -\\
Commanded base twist & 3 & - \\
Joint angles &  $12$ & -\\
Joint speeds & $12$ & -\\
Previous action & $12$ & -\\
Cosine / sine of phase & 2 & -\\
Height map & $(21,15)$ & -\\
Foot force wrenches & $(4, 6)$ & -\\
Foot heights & $4$ & -\\
Base mass & $1$ & $-1.0, +3.0$ \si{\kilogram} \\
Static / dynamic friction & $2$ & $[0.2, 1.2]$\\
Coefficient of restitution & $1$ & $[0.0, 0.3]$ \\
Joint stiffness & $12$ & $\pm 10\%$ \\
Joint damping & $12$ & $\pm 10\%$ \\
Joint friction & $12$ & $[0.0, 0.05]$ \\
\hline
\end{tabular}
\vspace{0.05cm}
\label{tab:state_qped}
\end{table}

\begin{table}[htb]
\centering
\caption{
\footnotesize{
State-based reward terms used for quadruped expert policy training.
}}
\begin{tabular}{lc}
\hline
Term & Weight \\
\hline
    Commanded $x,y$-velocity tracking reward & $1.5$ \\
    Commanded yaw rate tracking reward & $0.75$ \\
    Desired gait reward & $0.05$ \\
    Desired gait foot height reward & $0.2$ \\
    Base $z$-velocity penalty & $-2.0$ \\
    Base angular velocity penalty & $-0.05$ \\
    Base orientation penalty & $-4.0$ \\
    Deviation from default hip joint angles penalty & $-0.25$ \\
    Joint torque penalty & \num{-2.0e-4} \\
    Joint acceleration penalty & \num{-2.5e-7} \\
    Action rate penalty & $-0.01$ \\
\hline
\end{tabular}
\label{tab:reward_qped}
\end{table}

\begin{table}[htb]
\caption{\footnotesize{Observation space for the quadruped.}}
\centering
\begin{tabular}{lc}
\hline
Observation & Dimension \\
\hline
Base linear velocity (local frame) & 3 \\
Base angular velocity (local frame) & 3 \\
Projected gravity vector & 3\\
Joint angles & 12 \\
Joint speeds & 12 \\
Cosine and sine of phase & 2 \\
Height map & $(21, 15)$ \\
\hline
\end{tabular}
\vspace{0.05cm}
\label{tab:obs_qped}
\end{table}

\begin{table}[htb]
\caption{\footnotesize{World model architectural parameters for the quadruped.}}
\centering
\begin{tabular}{lc}
\hline
Parameter & Value \\
\hline
Embedding dimension & 64 \\
Proprioceptive observations MLP hidden dims & 128, 128 \\
CNN kernel size & 3\\
CNN strides & 2, 2, 2 \\
CNN features & 8, 16, 32 \\
All transformers MLP hidden size & 256 \\
Dynamics transformer layers & 2 \\
Dynamics transformer heads & 8 \\
Reward transformer layers & 1 \\
Reward transformer heads & 1 \\
Value transformer layers & 1 \\
Value transformer heads & 1 \\
Base policy transformer layers & 4 \\
Base policy transformer heads & 8 \\
\hline
\end{tabular}
\vspace{0.05cm}
\label{tab:model_qped}
\end{table}
\begin{table}[htb]
\caption{\footnotesize{MPPI parameters for the quadruped.}}
\centering
\begin{tabular}{lc}
\hline
Parameter & Value \\
\hline
Candidate actions batch size & 450 \\
Noised base policy actions batch size & 22 \\
Solver iterations & 8\\
Initial action standard deviation & 2.0 \\
Minimum action standard deviation & 0.05 \\
Base policy action standard deviation & 0.05 \\
Elites & 64 \\
Temperature & 0.25 \\
Momentum & 0.0 \\
Discount & 0.99 \\
\hline
\end{tabular}
\vspace{0.05cm}
\label{tab:mppi_qped}
\end{table}

\textbf{World Model Pretraining.}
We pretrain the world model for two epochs over the full simulation dataset $\simdataset$. Using a batch size of 512, this corresponds to approximately \num{3.69e5} gradient update steps. Optimization is performed with Adam using an initial learning rate of \num{2e-4}, which is annealed to \num{1e-4} via a cosine decay schedule, with a linear warmup over the first 10{,}000 steps. We apply data augmentation by injecting zero-mean Gaussian noise into the input observations (both proprioceptive and height map) during training. Pretraining requires approximately 28 hours on a single NVIDIA RTX 4500 Ada GPU.

\textbf{Hardware Deployment.} We use MPPI as implemented in TD-MPC \cite{hansen2022temporal} with the hyperparameters listed in \cref{tab:mppi_qped}. To encourage straight-line locomotion during quadruped experiments, we compute commanded base velocities $g_t$ from the robot’s current base pose using a PD controller on position, and provide these commands to the world model.

\textbf{Detailed Results.}
\Cref{tab:qpedallresults} reports detailed real-world quadruped locomotion results on both tasks across commanded forward speeds. We report both success rate (successful trials out of five) and average forward progress (mean $\pm$ standard deviation) over all trials at each speed. The \emph{Pretrained model} corresponds to zero-shot deployment of the simulation-pretrained world model without any real-world finetuning. While this model occasionally achieves partial forward progress, it fails to complete the task reliably, highlighting the severity of the sim-to-real dynamics gap. The \emph{Single-step BC policy}, which serves as the initial policy for the RLPD and IQL baselines prior to finetuning, improves performance in some settings but remains inconsistent and rarely achieves full task completion.

After real-world finetuning (32.1 minutes of data for Foam and 35.7 minutes for Slippery Slope), \Method\ consistently achieves the highest success rates and forward progress across all tested speeds. In contrast, both IQL and RLPD exhibit limited improvement despite access to the same real-world data budget. In particular, RLPD destabilized the robot during adaptation on the Foam task and is therefore not reported for that condition.

\begin{table*}[tb]
    \centering
    \caption{
    \footnotesize{
    Real-world quadruped results for both tasks.
    Success is reported as successful trials out of five.
    Forward progress is reported as mean $\pm$ standard deviation (meters) across trials at each commanded speed.
    The \emph{Pretrained model} corresponds to zero-shot deployment of the simulation-trained world model.
    The \emph{Single-step BC policy} is the behavior cloning policy used to initialize IQL and RLPD prior to finetuning.
    \Method, IQL, and RLPD results reflect performance after real-world finetuning using 35.7 minutes (Slippery Slope) and 32.1 minutes (Foam) of data.
    RLPD results on the Foam task are not reported, as the method destabilized the robot prior to evaluation.
    }
    }
    \begin{tabular}{lccccccccccc}
    \hline
        ~ & Speed & \multicolumn{2}{c}{Pretrained model} & \multicolumn{2}{c}{Single-step BC policy} & \multicolumn{2}{c}{\Method\ (ours)} & \multicolumn{2}{c}{IQL} & \multicolumn{2}{c}{RLPD} \\ 
        ~ & \si{\meter\per\second} & Fwd. Prog. & Success & Fwd. Prog. & Success & Fwd. Prog. & Success & Fwd. Prog. & Success & Fwd. Prog. & Success \\
        \hline
        Slippery & 0.1 & 0.70 $\pm$ 0.56 & 0/5 & 1.49 $\pm$ 0.29 & 2/5 & \textbf{1.78 $\pm$ 0.08} & \textbf{4/5} & 0.00 $\pm$ 0.00 & 0/5 & 0.32 $\pm$ 0.01 & 0/5 \\
        Slope & 0.3 & 0.43 $\pm$ 0.13 & 0/5 & 1.43 $\pm$ 0.25 & 1/5 & \textbf{1.82 $\pm$ 0.00} & \textbf{5/5} & 0.39 $\pm$ 0.56 & 0/5 & 0.34 $\pm$ 0.05 & 0/5 \\
        ~ & 0.5 & 0.90 $\pm$ 0.40 & 0/5 & 0.62 $\pm$ 0.23 & 0/5 & \textbf{1.82 $\pm$ 0.00} & \textbf{5/5} & 0.46 $\pm$ 0.42 & 0/5 & 0.35 $\pm$ 0.01 & 0/5 \\
        \hline
        Foam & 0.2 & 2.98 $\pm$ 0.02 & 3/5 & 2.07 $\pm$ 1.01 & 1/5 & \textbf{3.00 $\pm$ 0.00} & \textbf{5/5} & 0.92 $\pm$ 0.48 & 1/5 & -- & -- \\
        ~ & 0.7 & 2.39 $\pm$ 0.71 & 2/5 & 1.54 $\pm$ 0.81 & 1/5 & \textbf{3.00 $\pm$ 0.00} & \textbf{5/5} & 2.25 $\pm$ 0.87 & 2/5 & -- & -- \\
        ~ & 1.2 & 1.70 $\pm$ 0.99 & 0/5 & 2.45 $\pm$ 0.47 & 2/5 & \textbf{3.00 $\pm$ 0.00} & \textbf{5/5} & 2.73 $\pm$ 0.34 & 3/5 & -- & -- \\
        \hline
    \end{tabular}
    \label{tab:qpedallresults}
\end{table*}

\subsection{Ablations Details}
\label{app:ablations}
Each ablation corresponds to a separately pretrained world model. Across all ablations, the same planning hyperparameters and evaluation environments are used, and each configuration is evaluated on an identical set of randomized environments. Next, we discuss platform specific details.

\textbf{Manipulation.} For both manipulation tasks, we evaluate success rates over the initial condition and domain randomization described in the environments from \cite{yin2026emergent}.

\textbf{Quadruped.}
The robot is commanded to walk forward at a specified target speed until episode termination. For each evaluated model, we instantiate environments covering all combinations of parameters listed in \cref{tab:ablations_qped}, resulting in 1080 distinct environments per model. All models are evaluated on the same fixed set of environments to ensure fair comparison. Each environment is run for a single episode, during which state-based reward is accumulated until termination. Episodes terminate either after 1000 simulation steps or upon failure, defined as body contact with the ground or violation of base orientation limits. Results are summarized in \cref{tab:ablations}, where we report the average accumulated reward per episode across all environments.

\begin{table}[ht]
    \centering
    \caption{
    \footnotesize{
    Parameter values used to construct quadruped ablation environments in simulation. Each environment is defined by a unique combination of commanded forward speed, ground friction coefficient, terrain type, and terrain difficulty. All combinations are evaluated for each model, yielding 1080 environments per model.
    }
    }
    \begin{tabular}{|c|c|c|c|}
        \hline
        ~ & ~ & ~ & Terrain \\
        Speed & Friction & Terrain & Difficulty \\ \hline
        0.2 & 0.2 & Boxes & 0.2 \\ 
        0.4 & 0.4 & Rough & 0.4 \\ 
        0.6 & 0.6 & Stairs Up & 0.6 \\ 
        0.8 & 0.8 & Stairs Down & 0.8 \\ 
        1.0 & 1 & Slope Up & 1 \\ 
        1.2 & 1.2 & Slope Down & ~ \\
        \hline
    \end{tabular}
    \label{tab:ablations_qped}
\end{table}

\end{document}